# League of Legends: Real-Time Result Prediction

1st Jailson B. S. Junior
*Systems and Computing Department*
*Federal University of Campina Grande*
Campina Grande, Brazil
jailson.junior@ccc.ufcg.edu.br

2nd Claudio E. C. Campelo
*Systems and Computing Department*
*Federal University of Campina Grande*
Campina Grande, Brazil
campelo@dsc.ufcg.edu.br

*Abstract*—This paper presents a study on the prediction of outcomes in matches of the electronic game League of Legends (LoL) using machine learning techniques. With the aim of exploring the ability to predict real-time results, considering different variables and stages of the match, we highlight the use of unpublished data as a fundamental part of this process. With the increasing popularity of LoL and the emergence of tournaments, betting related to the game has also emerged, making the investigation in this area even more relevant. A variety of models were evaluated and the results were encouraging. A model based on LightGBM showed the best performance, achieving an average accuracy of 81.62% in intermediate stages of the match when the percentage of elapsed time was between 60% and 80%. On the other hand, the Logistic Regression and Gradient Boosting models proved to be more effective in early stages of the game, with promising results. This study contributes to the field of machine learning applied to electronic games, providing valuable insights into real-time prediction in League of Legends. The results obtained may be relevant for both players seeking to improve their strategies and the betting industry related to the game.

## I. Introduction

League of Legends is one of the most popular video games in the world today. There are over 150 million registered players, with an average of over 100 million monthly active players and over 10 million daily [1]. League of Legends (LoL) is a Multiplayer Online Battle Arena (MOBA) game developed by Riot Games in 2009. In the game, two teams face off with the objective of destroying the structures of the opposing team, with the main target being a structure called the nexus. The teams are composed of five players, each controlling a champion/character, and they battle on a map with three lanes and a jungle, with each player having a specific role in the match. There are many important objectives and variables in the game that directly impact victory, from destroying towers to killing dragons across the map, thereby generating advantages for the team in pursuit of victory.

With the game's fame and the growth of eSports [2, 3], the League of Legends World Championship, also known as Worlds, emerged. The event takes place annually and brings together hundreds of players from around the world, with the tournament being held in different cities every year. The first tournament was organized by Riot Games in 2011 and had a prize pool of US$100,000 [4]. Currently, the tournament is in its 12th edition, which took place in September 2022, with a total prize pool of US$2,225,000 [5].

With the popularity of tournaments and the associated prize money, betting and prediction pools have emerged within the community and even by Riot Games themselves. When Worlds begins, Riot Games creates a prediction pool for League of Legends players, where those who accurately predict the match outcomes receive in-game rewards. Additionally, there are betting websites that include eSports categories and allow users to bet on League of Legends matches. Websites like Betway and Rivalry are examples of platforms where you can place bets on LoL tournaments.

Predicting the outcomes of a LoL match can be quite challenging. Many factors can affect the final result, such as objective completions, ranging from jungle objectives like dragons and the Rift Herald to major objectives like tower destruction and inhibitors. The advantage one team has over the other in terms of gold (the in-game currency), items, or champion level is also taken into account. All these variables can change throughout the match, making it an even greater challenge to predict the final outcome.

This paper presents a machine learning-based approach to real-time prediction of LoL match results with the highest possible accuracy. Emphasis is placed on using recent and unique data as a fundamental part of this process, providing a comprehensive and up-to-date analysis of the game. Additionally, it aims to identify the best technique or combination thereof, considering the use of these distinct data sources. The effectiveness of the approach is also evaluated at different moments during the match, ensuring a dynamic and up-to-date approach to result prediction in the context of League of Legends.

The remainder of this paper is structured as follows. Section II presents background information. Then Section III discusses the related work. Following this, the data and methods employed are described in Section IV. Afterwards, Section V discusses the results. Finally, Section VI concludes the paper and points to future directions.

## II. Background

In this section, we provide a background that serves as a basis for understanding the analyses and models developed in this study. Firstly, we present an explanation of the MOBA (Multiplayer Online Battle Arena) genre, and subsequently, we delve into the understanding of the game League of Legends, which is the central focus of this work.

### A. MOBA Genre: Multiplayer Online Battle Arena

The MOBA genre, or Multiplayer Online Battle Arena, is one of the most popular genres in the competitive video game scene. MOBA games stand out for their combination of real-time strategy elements and intense action, providing a dynamic and challenging experience for players. In this genre, matches are contested between two teams, each composed of a group of players who control a single character, known as a hero or champion. Each hero has unique abilities and plays a specific role in the team, such as attack, defense, or support.

The central objective of MOBA games is to destroy the enemy base while defending one's own. The bases are protected by defensive towers and computer-controlled units that try to prevent the opposing team's advance. During the match, players must work together, plan strategies, coordinate attacks, and make tactical decisions to gain advantages and overcome opponents. Effective communication and synchronization among team members are crucial to success in the game.

Additionally, MOBA games feature elements such as character progression throughout the match, resource gathering, map objectives, and strategic decision-making. These elements contribute to the strategic depth and complexity of matches, making the MOBA genre a highly competitive and engaging form of gameplay.

### B. League of Legends

Among the numerous titles in the MOBA genre, League of Legends (LoL) stands out as one of the most popular and influential electronic games today. Developed by Riot Games, LoL was launched in 2009 and has garnered a massive player base worldwide.

In League of Legends, two teams composed of five players each face off on a strategically divided map with three main lanes (top, middle, and bottom) and an area called the jungle. Each player takes control of a champion with unique abilities, responsible for performing a specific role within the team. Champions are selected at the beginning of the match and have various play styles, such as fighters, mages, marksmen, tanks, and supports.

The main objective in LoL is to destroy the enemy Nexus, a structure located in the opposing team's base. To achieve this goal, players must advance through the lanes, eliminate computer-controlled minions, destroy defensive towers, and neutral objectives spread across the map, such as dragons and the Baron Nashor. These accomplishments provide resources and strategic advantages, such as gold, experience, and buffs, which can be used to strengthen the team and increase the chances of victory.

League of Legends stands out for its variety of champions, team strategies, ability combinations, and strategic complexity. Each match is unique, requiring quick adaptation, tactical decision-making, and efficient teamwork. The depth of the game and the constantly evolving meta game (dominant strategies) contribute to its longevity and continuous popularity in the eSports scene.

## III. RELATED WORK

This section analyzes research related to the prediction of outcomes in electronic games, with a focus on MOBA and League of Legends. These studies explore different ways to predict who will win the matches and understand what affects the success of teams. This review gives us an idea of what researchers have been studying in this field.

### A. Prediction of Results in MOBA Games

In the area of predicting results in MOBA games, specifically in the context of DOTA 2, studies have been conducted that contribute to the advancement of this field. Ke et al. [6] propose an innovative approach to identify and define team fights as crucial events during DOTA 2 matches. The main objective of this study is to explore the potential of team fight information in real-time prediction of match results. By using different recurrent neural network models, the study achieved an accuracy of over 70% when considering all team fight information up to the first 32 minutes of each match.

Another study conducted in the context of DOTA 2 investigated the use of machine learning models, from supervised linear regression to deep learning models such as Neural Networks and Long Short-Term Memory (LSTM) [7]. The results obtained demonstrated that deep learning models achieved high accuracy rates, with averages of 82% for linear regression and up to 93% for LSTM models. Furthermore, analyses were conducted considering different future prediction stages and temporal correlation, revealing the importance of these aspects in obtaining more accurate results.

In a different context, in the game Heroes of the Storm (HOTS), Swidler [8] explored the possibility of predicting match outcomes. Players have the option to upload replay files to a website called HOTSLogs [9], which analyzes the games and provides relevant statistics. Although Blizzard Entertainment, the game developer, does not provide a match history for download, HOTSLogs offers a way to obtain a summary of the last 30 days of games. This data was used to develop a model that accurately predicts over 62% of matches and estimates the probability of a team's victory in a given game.

### B. Prediction of League of Legends Results

The articles analyzed in this section address the prediction of League of Legends match results using different machine learning approaches. In common, all studies aim to identify patterns and variables that may influence match outcomes in order to develop accurate prediction models.

Some works approach the prediction of League of Legends match results using neural networks [10, 11]. They explore different sets of features such as Dragon, Level, Rift Heralds, Towers, gold, kills, assists, destroyed towers, and drakes. These studies demonstrate the capability of neural networks to achieve high prediction accuracy rates, ranging from 75.1% to 93.75%.

In contrast, other studies utilize decision tree algorithms to predict outcomes [12,13]. These studies consider features such as duration, kills, towers, and gold. Both studies obtained promising results, with average accuracies of 80% and accuracy rates ranging from 68.33% to 85.17%.

On the other hand, Silva et al. [14] compare different types of Recurrent Neural Networks (RNNs) in result prediction. A simple RNN achieved higher accuracy and was selected for further experiments with different time intervals. The accuracy ranged from 63.91% to 83.54% depending on the considered time interval. The study suggests the use of RNNs to analyze power spikes and make strategic decisions in the game.

Do et al. [15] highlight the importance of champion selection and player experience in result prediction. A machine learning model using Deep Neural Network (DNN) achieved an accuracy of 75.1% when considering the player's experience with the champion as one of the features. The study emphasizes the need for fair matchmaking systems that take into account players' champion selection.

In summary, the existing approaches share the goal of predicting League of Legends match results but differ in their approaches and considered features. Neural networks and decision tree algorithms have shown promise, while the analysis of time intervals and player experience with the champion have also been addressed. These studies provide valuable insights for the gaming community and developers, providing a solid foundation for future work in the field of result prediction in competitive games like League of Legends.

Table I presents a comparison of related works, highlighting the techniques used and the data considered in each study.

TABLE I
COMPARATIVE TABLE OF RELATED WORKS (TECHNIQUES AND DATA)

| Ref | Techniques | | | | | | Data | |
|---|---|---|---|---|---|---|---|---|
| | DNN | Decision Trees | RNN | Logistic Regression | RF | Gradient Boosting | Pre-Game | In-Game |
| [10] | X | | | | | | | X |
| [12] | | X | | | X | | | X |
| [11] | | | | X | X | X | X | X |
| [14] | | | X | | | | X | X |
| [15] | X | | | | | | X | X |
| [13] | | X | | | | | | X |
| [16] | X | | | | | | | X |

## IV. METHODOLOGY

This Methodology section describes the approach adopted in this research for predicting League of Legends games. We present the dataset used, the applied machine learning techniques, and the evaluation metrics employed. The objective is to explore different approaches and provide accurate predictions to enhance the understanding and performance in this competitive gaming context.

### A. Data Collection

The data was collected using the official API from Riot Games [17], the developer of the game League of Legends. The API provides access to a variety of game-related information, such as match statistics, player details, results, and other relevant data for the study. Through the API, it was possible to obtain up-to-date and accurate data for analysis and modeling.

The data collection process involved the use of the *riotwatcher* library [18], a Python library specifically developed to interact with the official Riot Games API. This library facilitated the sending of requests and the manipulation of the data returned by the API, streamlining the collection process.

It is important to highlight that the data collection through the Riot Games API, with the assistance of the *riotwatcher* library, ensured the reliability and integrity of the data, as the information was obtained directly from the official source of the game. Additionally, all privacy policies and terms of use of the API were strictly followed during the collection process.

### B. Dataset

The dataset used in this study represents a representative sample of ranked matches in League of Legends, encompassing players of different skill levels, ranging from Iron to Diamond elo. The 64,556 collected matches provide a comprehensive and diverse view of the competitive landscape during the period from January 12, 2023, to May 18, 2023. [19]

Each recorded match in the dataset contains a wealth of valuable information for analysis and modeling. Among the available variables, we have information about champion kills, including the number of kills performed by the blue and red teams, the first kill that occurred in each match, as well as the players' performance in securing kills and assists.

Additionally, we have data on the achievement of important objectives during matches, such as the elimination of dragons, towers, and inhibitors. These strategic elements are crucial in determining a team's advantage and their ability to control the game map.

Other relevant variables include the total gold acquired by each team, the number of minions slain both in lanes and in the jungle, the average level of players, and many other metrics that allow us to better understand the dynamics of matches and the factors that can influence the final result.

By exploring this diverse dataset, the goal is to identify patterns, trends, and correlations among the variables in order to develop more accurate and robust prediction models for predicting the outcomes of League of Legends matches. This detailed analysis of the dataset is essential for understanding the game and developing more effective strategies.

### C. Data Preprocessing

During the data preprocessing process, several approaches were adopted to ensure the quality and consistency of the dataset. One of the initial steps involved converting time values, expressed in milliseconds, to minutes. This transformation allowed for a better understanding of the match duration and facilitated the analysis of events at different moments of the game.

The dataset was then divided into four scenarios, representing different percentages of the total match duration: 20% PET (Percent Elapsed Time), 40% PET, 60% PET, and 80% PET. This strategic division, using the concept of Percent Elapsed Time (PET), was carried out to examine the performance of the model when dealing with varying amounts of information available about the game. While intuition leads us to expect that the model will make better predictions as more information is provided, it is important to highlight that the performance of the model does not always exhibit a linear growth as the amount of information increases. This analytical approach allows us to explore how the model behaves in different temporal contexts.

To ensure the reliability and of the data, measures were taken to address potential issues with data quality. As part of this process, matches with a duration of less than 5 minutes were excluded from the dataset. This decision was made to eliminate instances that are likely to be remakes rather than actual matches.

Another important step was the transformation of boolean variables into binary numeric values, assigning the value *0* for "false" and the value *1* for "true". This approach simplifies the representation of the data and allows for the direct application of modeling and numerical analysis techniques.

Additionally, redundant or opposing variables were removed. For example, the variables *redWin* and *redFirstBlood* were removed as they provide similar information to the remaining variables and do not add additional value to the analysis. This removal of redundant variables resulted in a more concise dataset focused on the most relevant information for the study.

*D. Exploratory Data Analysis*

In this subsection, we will perform an exploratory data analysis to understand the general characteristics, identify patterns and trends, and extract relevant insights for our analysis. We will use visualization techniques and descriptive statistics to explore the data in detail and gain a comprehensive view of the dataset.

We now present a concise overview of the key findings and observations derived from our Exploratory Data Analysis.

During the exploratory data analysis, it was observed that the win ratio is 50.5% for the blue team and 49.5% for the red team. This indicates that there is no significant imbalance between the teams, with a relatively balanced outcome.

Regarding the duration of the matches, the longest one recorded in the dataset is 67.35 minutes, while the median duration of the matches is 30.01 minutes. This suggests that the majority of matches have a duration around this mark.

Fig. 1 shows the distribution of match durations in minutes for both the complete match (100%) and each of the portions considered in predictive analysis (20%, 40%, 60%, 80%). These values provide insights into the progression of game time throughout the matches.

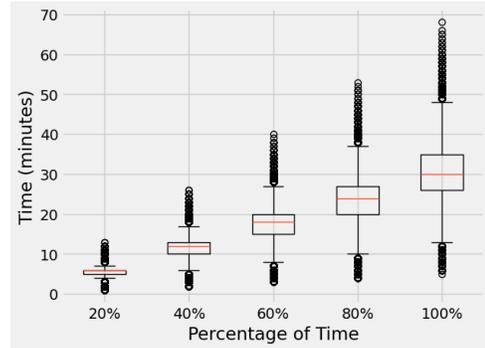

Fig. 1. Variation of Match Durations in LoL Across Different Time Intervals

We examined the correlations between the variables and the victory of the blue team at different time intervals. We used the Pearson correlation coefficient to assess the linear relationships between these variables and the victory of the blue team. The results of these correlations are summarized in Tables II, III, IV, and V.

TABLE II
CORRELATIONS BETWEEN VARIABLES AND BLUE TEAM VICTORY - 20% OF MATCHES

| Positive Variables | Correlation | Negative Variables | Correlation |
|---|---|---|---|
| blueFirstBlood | 0.14 | redFirstBlood | -0.15 |
| blueChampionKill | 0.10 | redChampionKi | -0.12 |
| blueTotalGold | 0.04 | redTotalGold | -0.07 |

TABLE III
CORRELATIONS BETWEEN VARIABLES AND BLUE TEAM VICTORY - 40% OF MATCHES

| Positive Variables | Correlation | Negative Variables | Correlation |
|---|---|---|---|
| blueChampionKill | 0.16 | redChampionKill | -0.18 |
| blueFirstBlood | 0.16 | redDragonKill | -0.17 |
| blueDragonKill | 0.16 | redFirstBlood | -0.16 |

TABLE IV
CORRELATIONS BETWEEN VARIABLES AND BLUE TEAM VICTORY - 60% OF MATCHES

| Positive Variables | Correlation | Negative Variables | Correlation |
|---|---|---|---|
| blueDragonKill | 0.29 | redDragonKill | -0.31 |
| blueChampionKill | 0.22 | redChampionKill | -0.24 |
| blueFirstBlood | 0.16 | redTowerKill | -0.19 |

When analyzing Tables II, III, IV, and V, we can notice that the correlations between the variables and the victory of the blue team vary according to the elapsed time in the match. In the early stages, as shown in Table II, the performance of the blue team's champions in kills and the achievement of the first blood are relevant. As the match progresses, Tables III, IV, and V demonstrate that the achievement of Dragons and the destruction of towers by the blue team also become significant. On the other hand, the performance of the red team in kills, the achievement of Dragons, and the destruction of towers by the red team have a negative impact on the victory of the blue team. These insights help us understand which aspects are more determinant for the success of the blue team at different moments of the game.

TABLE V
CORRELATIONS BETWEEN VARIABLES AND BLUE TEAM VICTORY - 80% OF MATCHES

| Positive Variables | Correlation | Negative Variables | Correlation |
|---|---|---|---|
| blueDragonKill | 0.43 | redDragonKill | -0.45 |
| blueChampionKill | 0.29 | redTowerKill | -0.38 |
| blueTowerKill | 0.27 | redChampionKill | -0.31 |

*E. Feature Selection*

When analyzing the correlation between the variables and the victory of the blue team, we observed that different variables showed different levels of correlation in different scenarios of the matches. Overall, the variables *blueChampionKill*, *blueDragonKill*, and *blueFirstBlood* showed a stronger positive correlation with the victory of the blue team in almost all the analyzed scenarios. This suggests that having a higher number of champion kills, objectives related to dragons, and securing the first kill are associated with a higher probability of victory for the blue team.

However, it is important to highlight that correlation does not imply direct causality. Although these variables show a significant correlation, other factors may influence the outcome of the match. Therefore, it is necessary to consider these variables in conjunction with other relevant information to build more accurate prediction models.

Additionally, during the analysis of the different scenarios, we observed some differences in the correlations of the variables. For example, in the 40% PET scenario of the matches, the initial kill (*firstBlood*), the number of kills, and the achievement of dragons showed equal correlations with the victory of the blue team, indicating the importance of these events in the team's performance.

In the 60% PET scenario of the matches, the variables related to champion kills, dragons, and the first kill showed more significant correlations with the victory of the blue team. This may suggest that these objectives are crucial for the team's success in this stage of the game.

In the 80% PET scenario of the matches, we observed a strengthening of the correlations between champion kills, dragons, and towers conquered by the blue team. These results may indicate that in more advanced matches, these aspects become even more determinative for victory.

Therefore, considering these correlation analyses in different scenarios, we can conclude that certain variables have a more significant relationship with the victory of the blue team in different stages of the matches.

Table VI presents the variables that were taken into account in the construction of prediction models based on the correlation analyses discussed earlier.

TABLE VI
VARIABLES USED FOR TRAINING.

| Variable | Description |
|---|---|
| blueChampionKill | Champion kills by blue team |
| blueFirstBlood | Blue team's first kill indicator |
| blueDragonKill | Dragons killed by blue team |
| blueDragonHextechKill | Hextech dragons killed by blue team |
| blueDragonChemtechKill | Chemtech dragons killed by blue team |
| blueDragonFireKill | Fire dragons killed by blue team |
| blueDragonAirKill | Air dragons killed by blue team |
| blueDragonEarthKill | Earth dragons killed by blue team |
| blueDragonWaterKill | Water dragons killed by blue team |
| blueDragonElderKill | Elder dragons killed by blue team |
| blueRiftHeraldKill | Blue team's Rift Herald kill indicator |
| blueBaronKill | Blue team's Baron Nashor kill indicator |
| blueTowerKill | Towers destroyed by blue team |
| blueInhibitorKill | Inhibitors destroyed by blue team |
| blueTotalGold | Total gold accumulated by blue team |
| blueMinionsKilled | Minions killed by blue team |
| blueJungleMinionsKilled | Jungle monsters killed by blue team |
| blueAvgPlayerLevel | Average level of blue team players |
| redChampionKill | Champion kills by red team |
| redDragonKill | Dragons killed by red team |
| redDragonHextechKill | Hextech dragons killed by red team |
| redDragonChemtechKill | Chemtech dragons killed by red team |
| redDragonFireKill | Fire dragons killed by red team |
| redDragonAirKill | Air dragons killed by red team |
| redDragonEarthKill | Earth dragons killed by red team |
| redDragonWaterKill | Water dragons killed by red team |
| redDragonElderKill | Elder dragons killed by red team |
| redRiftHeraldKill | Red team's Rift Herald kill indicator |
| redBaronKill | Red team's Baron Nashor kill indicator |
| redTowerKill | Towers destroyed by red team |
| redInhibitorKill | Inhibitors destroyed by red team |
| redTotalGold | Total gold accumulated by red team |
| redMinionsKilled | Minions killed by red team |
| redJungleMinionsKilled | Jungle monsters killed by red team |
| redAvgPlayerLevel | Average level of red team players |

The variables listed in Table VI represent the aspects identified as having a significant correlation with the victory of the blue team at different stages of the matches.

This information is valuable for the construction of prediction models that take into account these specific characteristics to predict the outcome of League of Legends matches with greater accuracy.

*F. Construction and Evaluation of Prediction Models*

In this subsection, we describe the methodology adopted for the construction and evaluation of prediction models for League of Legends matches, taking into consideration the percentage of elapsed time in the matches. In addition to the machine learning algorithms mentioned earlier, we considered this new variable as an important attribute to enhance the accuracy of the models.

Several machine learning algorithms were selected for this study, including Logistic Regression, Random Forest, Naive Bayes, scikit-learn's Gradient Boosting, XGBoost, LightGBM, Neural Network, and Bagging. This choice was motivated by these algorithms' capability to handle classification problems and their common usage in prediction studies.

The evaluation of the models was conducted using the cross-validation technique, an approach that provides robust evaluation considering the percentage of elapsed time as part of the input data. This method allows us to capture patterns and behaviors over time, contributing to the improvement of the predictive capability of the models.

During the model evaluation, essential metrics such as accuracy, precision, recall, F1-score, and AUC-ROC were applied. Furthermore, we considered the models' performance and predictive ability while taking into account the percentage of elapsed time in the matches.

As part of model optimization, we chose to employ the hyperparameter search technique known as Random Search on the Random Forest and LightGBM algorithms. This approach enabled us to efficiently explore a wide range of hyperparameter combinations, aiming to identify more promising configurations for the models' performance.

When comparing the results obtained from various models, we did not limit ourselves to the analysis of accuracy and traditional metrics alone. We also considered the models' ability to capture and interpret variations over time. This approach enables us to identify the most suitable model for predicting the outcomes of League of Legends matches, considering the impact of the percentage of elapsed time.

By including the percentage of elapsed time as an additional attribute, we were able to obtain more precise and relevant insights for predicting match outcomes. The combination of machine learning algorithms, cross-validation, evaluation metrics, and the inclusion of the percentage of elapsed time enriches the construction of effective and reliable prediction models.

## V. RESULTS AND DISCUSSION

During the analysis of the correlations between the variables and the victory of the blue team, we observed that these correlations vary throughout the duration of the match. In the early stages, the kills made by the champions of the blue team and the achievement of first blood were strongly correlated with victory. As the match progressed, the acquisition of Dragons and the destruction of towers by the blue team also showed a positive correlation with the outcome.

Regarding the feature importance of the random forest model, we observed that it varies over time in the match. Specifically, at 20% of the elapsed time, the most important variable for predicting the victory of the blue team was *redTotalGold*. This means that the total gold obtained by the red team played a crucial role in determining the outcome in this early phase of the match. This observation is evidenced in Fig. 2, where we can see the graph of the top 12 features with the highest importances in this scenario.

In the 40% PET, 60% PET, and 80% PET scenarios, the importance of the features underwent changes. At 40% PET scenario, the variable "blueTotalGold" became the most relevant, as illustrated in Fig. 3. This indicates that the total gold acquired by the blue team was a determining factor for their victory in this intermediate phase of the match. When PET=60%, it could be observed that the variable "blueTotalGold" remained the most important, as shown in

Fig. 4, reinforcing its relevance in this stage of the game. For PET=80%, the variable "blueChampionKill" held the highest importance, as shown in Fig. 5, emphasizing the significance of champion kills executed by the blue team in this stage of the match.

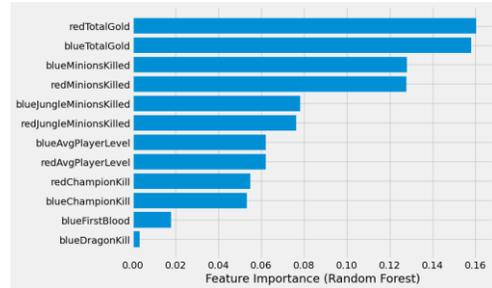

Fig. 2. Feature Importance Graph at 20% Time

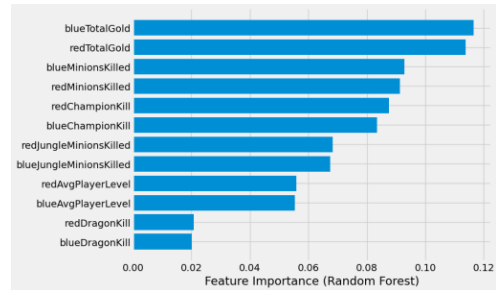

Fig. 3. Feature Importance Graph at 40% Time

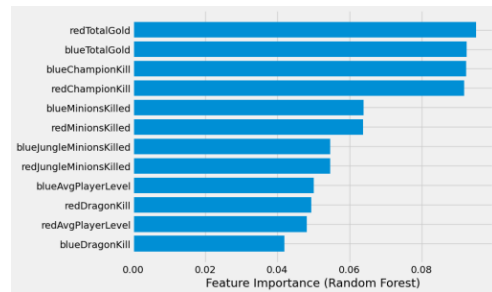

Fig. 4. Feature Importance Graph at 60% Time

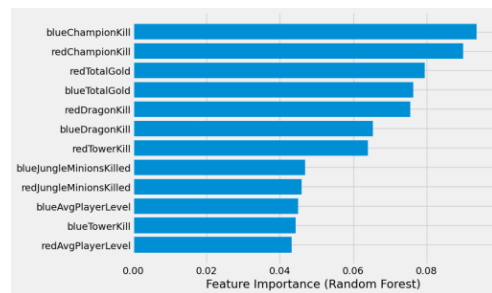

Fig. 5. Feature Importance Graph at 80% Time

These findings highlight how the relevant features for the victory of the blue team can vary over time. Therefore, it is essential to consider the different stages of the match when building prediction models and developing strategies to succeed at different moments of the game. Additionally, when comparing the results of correlations with the feature importance obtained through Random Forest, significant differences can be noticed. While correlations provide information about the direct relationship between variables and victory, feature importance indicates the degree of contribution of each variable to the accuracy of the prediction model.

It is interesting to note that, in some cases, a variable may have a strong correlation with the victory of the blue team, but relatively low importance according to Random Forest. This can be explained by the fact that Random Forest evaluates the importance of a variable considering not only its direct relationship with the target variable but also its interaction with other variables and its ability to reduce model uncertainty.

Therefore, it is important to consider both correlations and feature importance when interpreting the results. While correlations help identify the direct relationships between variables and victory, feature importance provides insights into the relative impact of each variable on the overall performance of the prediction model. This combination of information allows us to obtain a more comprehensive and accurate understanding of the factors that influence the outcome of League of Legends matches.

Based on the results obtained from applying the constructed prediction models, we evaluated the performance of each of them considering the percentage of elapsed time in League of Legends matches. The evaluation metrics used were accuracy, precision, recall, F1-score, and AUC-ROC.

Table VII presents the evaluation metrics for each model at different PET.

In the analysis of accuracy metrics among models across different time intervals, we can observe interesting trends. When considering 20% of the elapsed time, the standout models were Logistic Regression, followed by Gradient Boosting and XGBoost. In the 40% interval, we had LightGBM leading, followed by Logistic Regression and again XGBoost.

Moving on to the 60% elapsed time mark, LightGBM achieved the highest performance, followed by Random Forest and Gradient Boosting. Finally, reaching 80% of the elapsed time, LightGBM once again demonstrated the best accuracy, followed by Random Forest and XGBoost.

This analysis showcases the variation in model performance throughout the course of the match. LightGBM consistently excelled across different intervals, highlighting its adaptability to various phases of the game. It's worth noting that the success of each model could depend on specific match factors, and considering the percentage of elapsed time as an additional attribute could be beneficial for the model learning process.

Table VIII presents the accuracy rates achieved through the application of the hyperparameter search technique known as Random Search on the LightGBM and Random Forest models.

TABLE VII
AVERAGE EVALUATION METRICS OF MODELS - ELAPSED TIME PERCENTAGE IN MATCHES

| Model | Elapsed Time Percentage | Accuracy | Precision | Recall | F1-score | AUC-ROC |
|---|---|---|---|---|---|---|
| Random Forest | 20% | 0.6074 | 0.6132 | 0.6014 | 0.6072 | 0.6577 |
|  | 40% | 0.6989 | 0.7064 | 0.6900 | 0.6981 | 0.7841 |
|  | 60% | 0.7781 | 0.7837 | 0.7738 | 0.7787 | 0.8710 |
|  | 80% | 0.8533 | 0.8569 | 0.8516 | 0.8542 | 0.9375 |
| Logistic Regression | 20% | 0.6228 | 0.6244 | 0.6342 | 0.6292 | 0.6589 |
|  | 40% | 0.7021 | 0.7007 | 0.7149 | 0.7077 | 0.7505 |
|  | 60% | 0.7722 | 0.7730 | 0.7768 | 0.7749 | 0.8368 |
|  | 80% | 0.8476 | 0.8480 | 0.8505 | 0.8492 | 0.9195 |
| Naive Bayes | 20% | 0.5801 | 0.5698 | 0.6868 | 0.6224 | 0.6205 |
|  | 40% | 0.6366 | 0.6309 | 0.6747 | 0.6520 | 0.6848 |
|  | 60% | 0.7256 | 0.7247 | 0.7356 | 0.7301 | 0.7885 |
|  | 80% | 0.8141 | 0.8136 | 0.8194 | 0.8165 | 0.8817 |
| Gradient Boosting | 20% | 0.6181 | 0.6208 | 0.6252 | 0.6229 | 0.6756 |
|  | 40% | 0.7004 | 0.7050 | 0.6984 | 0.7017 | 0.7872 |
|  | 60% | 0.7765 | 0.7773 | 0.7807 | 0.7790 | 0.8713 |
|  | 80% | 0.8501 | 0.8509 | 0.8522 | 0.8516 | 0.9361 |
| XGBoost | 20% | 0.6179 | 0.6200 | 0.6256 | 0.6228 | 0.6178 |
|  | 40% | 0.7014 | 0.7043 | 0.7029 | 0.7036 | 0.7014 |
|  | 60% | 0.7691 | 0.7730 | 0.7674 | 0.7702 | 0.7691 |
|  | 80% | 0.8516 | 0.8557 | 0.8487 | 0.8522 | 0.8516 |
| LightGBM | 20% | 0.6164 | 0.6186 | 0.6261 | 0.6222 | 0.6744 |
|  | 40% | 0.7036 | 0.7074 | 0.7035 | 0.7055 | 0.7913 |
|  | 60% | 0.7788 | 0.7798 | 0.7827 | 0.7812 | 0.8746 |
|  | 80% | 0.8542 | 0.8545 | 0.8571 | 0.8558 | 0.9409 |
| Neural Network | 20% | 0.5614 | 0.6136 | 0.6646 | 0.5504 | 0.6504 |
|  | 40% | 0.5772 | 0.5577 | 0.9480 | 0.6967 | 0.6975 |
|  | 60% | 0.7155 | 0.7924 | 0.6820 | 0.6802 | 0.8358 |
|  | 80% | 0.7337 | 0.8605 | 0.6786 | 0.7095 | 0.8753 |
| Bagging | 20% | 0.5835 | 0.6023 | 0.5139 | 0.5545 | 0.6204 |
|  | 40% | 0.6726 | 0.7003 | 0.6140 | 0.6543 | 0.7468 |
|  | 60% | 0.7523 | 0.7821 | 0.7059 | 0.7420 | 0.8428 |
|  | 80% | 0.8361 | 0.8570 | 0.8106 | 0.8331 | 0.9195 |

TABLE VIII
ACCURACY USING RANDOM SEARCH

| Model | 20% | 40% | 60% | 80% |
|---|---|---|---|---|
| LightGBM | 0.6189 | 0.7036 | 0.7786 | 0.8548 |
| Random Forest | 0.6180 | 0.7034 | 0.7799 | 0.8543 |

This efficient approach explored various combinations of hyperparameters, aiming to optimize the most promising models through the Random Search method. The comparison between optimized and non-optimized versions revealed noticeable improvements in accuracy, validating the effectiveness of Random Search in enhancing predictive performance. In summary, the utilization of Random Search on the most promising models led to significant performance enhancements, underscoring its importance in optimizing models and obtaining more reliable results.

It is noteworthy that, overall, the models demonstrated enhanced performance when considering the percentage of elapsed time as an additional attribute in their learning processes.

Fig. 6 displays the average accuracy of the models at different elapsed time percentages.

We can observe a trend of increasing average accuracy of the models as the elapsed time percentage increases. This indicates that the information about the elapsed time in the matches is relevant for predicting the outcomes.

VI. CONCLUSIONS AND FUTURE WORK

Based on the results obtained and the conducted analyses, we can conclude that including the elapsed time percentage in League of Legends matches as an additional attribute has a significant impact on the performance of prediction models.

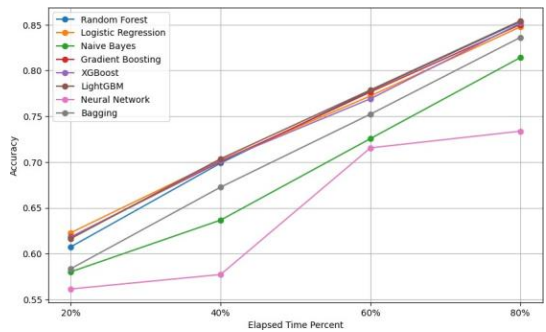

Fig. 6. Average Accuracy Across Elapsed Time Percent for Different Models.

The models showed overall improvement in their evaluation metrics when considering this temporal information.

The LightGBM model proved to be the most effective in terms of performance, especially when the elapsed time percentage was between 40% and 80%. However, in the early stages of the game (elapsed time percentage of 20%), other models such as Logistic Regression and Gradient Boosting had superior results. This indicates that different models may be more suitable depending on the stage of the match.

Considering an accuracy of 60%, we can state that there is potential for making predictions in games that have only passed 20% of the time. However, it is important to note that predicting outcomes in League of Legends matches still presents significant challenges due to the dynamic nature of the game. Therefore, it is necessary to consider other factors beyond model accuracy, such as the analysis of other performance metrics and understanding the limitations of the study.

The insights gained revealed the importance of different variables over time. Variables related to champion kills, dragons, and conquered towers had a greater impact on predictions in the early stages, while variables such as total gold and first blood became more relevant as the elapsed time increased. This game evolution highlights the importance of considering different stages of the matches when building prediction models.

Although the results were promising, there is still room for improvement. The dynamic and complex nature of League of Legends poses significant challenges for modeling and predicting outcomes. Therefore, it is crucial to consider other variables and additional information, as well as explore new approaches and datasets, in order to further enhance the prediction models.

An interesting perspective for future work is to address the challenge of predicting the outcomes of League of Legends matches as a time series. This approach could enable the capture of emerging patterns and trends in the game, considering the dynamic and ever-evolving nature of the matches. Additionally, a compelling suggestion for future work is the real-time guidance implementation. This implementation could provide predictions to players during the actual gameplay, enhancing their experience and enabling tests of the practical effectiveness of predictions in a real game scenario. As a complementary notion, the evaluation of real-time performance is also valuable. Investigating practical aspects of the model, such as latency and processing rate, in conjunction with real-time implementation, would yield significant insights into the challenges and requirements associated with real-time deployment. By exploring these three research directions, we aim to broaden the applicability and effectiveness of predictions within the ever-changing context of League of Legends.